\documentclass[runningheads]{llncs}

\usepackage{graphicx}
\usepackage{times}
\usepackage{epsfig}
\usepackage{graphicx}
\usepackage{amsmath}
\usepackage{amssymb}
\usepackage{multirow}
\usepackage{subcaption}
\usepackage{tikz} 
\usepackage[ruled,linesnumbered]{algorithm2e} 

\usepackage{xcolor}

\newcommand{\white}{\textcolor[rgb]{1.0,1.0,1.0}}
\newcommand{\blk}{\textcolor[rgb]{0.0,0.0,0.0}}
 
\newcommand{\tilo}{\textcolor[rgb]{0.3,0.3,1.0}} %
\newcommand{\nwc}{\textcolor[rgb]{0.0,0.0,0.0}} %https://www.overleaf.com/project/6093e927157d185797386ac5

\begin{document}

\title{\white{.}\vspace{-60pt}\\Label a Herd in Minutes:\\Individual Holstein-Friesian Cattle Identification\vspace{-10pt}\thanks{Jing Gao was supported by the China Scholarship Council. This work also benefitted from ground work carried out under the EPSRC grant EP/N510129/1 and the John Oldacre Foundation. Many thanks to Andrew Dowsey and Will Andrew for their contributions.}
% \thanks{This work is supported by xxxx.}
}

\titlerunning{Label a Herd in Minutes}

%
\author{Jing Gao \and Tilo Burghardt\ \and Neill W. Campbell}
\authorrunning{J. GAO et al.}

\institute{Department of Computer Science, University of Bristol, Bristol, United Kingdom \\
\email{\scriptsize{jing.gao@bristol.ac.uk, tilo@cs.bris.ac.uk, neill.campbell@bristol.ac.uk}}\vspace{-5pt}
}

%
% \author{LPLF submission no. 01}
% \institute{}
\maketitle  % typeset the header of the contribution

\vspace{-11pt}
\begin{abstract}
\nwc{We describe a practically evaluated approach for training visual cattle ID systems for a whole farm requiring only ten minutes of labelling effort.
In particular, for the task of automatic identification of individual Holstein-Friesians in real-world farm CCTV, we show that self-supervision, metric learning, cluster analysis, and active learning can complement each other to significantly reduce the annotation requirements usually needed to train cattle identification frameworks. Evaluating the approach on the test portion of the publicly available Cows2021 dataset, for training we use 23,350 frames across 435 single individual tracklets generated by automated oriented cattle detection and tracking in operational farm footage. Self-supervised metric learning is first employed to initialise a candidate identity space where each tracklet is considered a distinct entity. Grouping entities into equivalence classes representing cattle identities is then performed by automated merging via cluster analysis and active learning. Critically, we identify the inflection point at which automated choices cannot replicate improvements based on human intervention to reduce annotation to a minimum. Experimental results show that cluster analysis and a few minutes of labelling after automated self-supervision can improve the test identification accuracy of 153 identities to 92.44\% (ARI=0.93) from the 74.9\% (ARI=0.754) obtained by self-supervision only. These promising results indicate that a tailored combination of human and machine reasoning in visual cattle ID pipelines can be highly effective whilst requiring only minimal labelling effort. We provide all key source code and network weights with this paper for easy result reproduction.
% assume 30questions in 1min 
% 477-173 = 294 queries 294/30 = 10
}

\vspace{-3pt}
\end{abstract}

%====================================================================================
\vspace{-29pt}
\section{Introduction}
\vspace{-10pt}
$\bf{Background.}$ \nwc{Individual animal identification is mandatory~\cite{eu82097} in dairy farming and critical for managing aspects such as disease outbreaks and animal welfare. To date, invasive identification methods~\cite{awad2016classical} such as ear tags, tattoos, radio-frequency tags or branding are most often deployed. However, ethical considerations aside, these techniques cannot provide continuous location and ID information which still requires specialist tracking systems. For Holstein-Friesians, which constitute the most numerous and also highest milk-yielding~\cite{tadesse2003milk} cattle breed, contactless visual biometric (re)identification methods~\cite{kuhl2013,schneider2020similarity} using their characteristic black-and-white skin markings~\cite{hu2020cow,li2017automatic,andrew2016automatic,andrew2017visual,andrew2019visual,andrew2021visual} have become viable due to advances in deep learning. These approaches produce continuous coverage as long as cameras cover the whole farming area of interest. Similarly, face~\cite{yao2019cow}, muzzle~\cite{awad2016classical}, retina~\cite{awad2016classical}, or rear~\cite{qiao2019individual} biometrics may also be used in specific settings. However, modern biometric deep learning approaches that underpin systems for larger herds require significant amounts of identity-annotated visual data, demanding weeks of human annotation efforts~\cite{andrew2021visual}.
}
\ \\
$\bf{Conceptual}$ $\bf{Approach.}$ \nwc{To address this problem and reduce labelling requirements, the literature has recently fielded self-supervision methods~\cite{gao2021towards}, which learn by exploiting the internal structure of data. However, although superior to traditional unsupervised approaches~\cite{yu2017cross} the accuracy achieved with such systems still lags significantly behind benchmarks using supervised deep learning~\cite{andrew2021visual}. In response, here we advocate combining self-supervision, the analysis of the constructed identity space, and minimal active learning~\cite{settles2009active} to improve performance whilst limiting annotation requirements. Noting that research into visual cattle ID systems with reduced labelling is in its infancy~\cite{andrew2021visual,gao2021towards,vidal2021perspectives}, we propose a hybrid training approach~\cite{wu2018exploit,rizve2020defense,wang2018deep,wang2016cost} for learning cattle IDs from RGB videos. To  isolate the impact of different sources of training information, we follow a three-phase strategy:\vspace{-5pt} 
\begin{itemize}
    \item PHASE \#1: Exploit data-internal structure (via self-supervised metric learning).
    \item PHASE \#2: Exploit latent identity space structure (via cluster analysis).
    \item PHASE \#3: Utilise limited and targeted user input (via active learning).\vspace{-25pt}
\end{itemize}} \ \\

%--------------------------------------------------
\begin{figure}[h]
    \centering
    \vspace{-9pt}
    \includegraphics[width=1.03\textwidth, height=170px, %178
    trim= 10 10 0 0, clip]{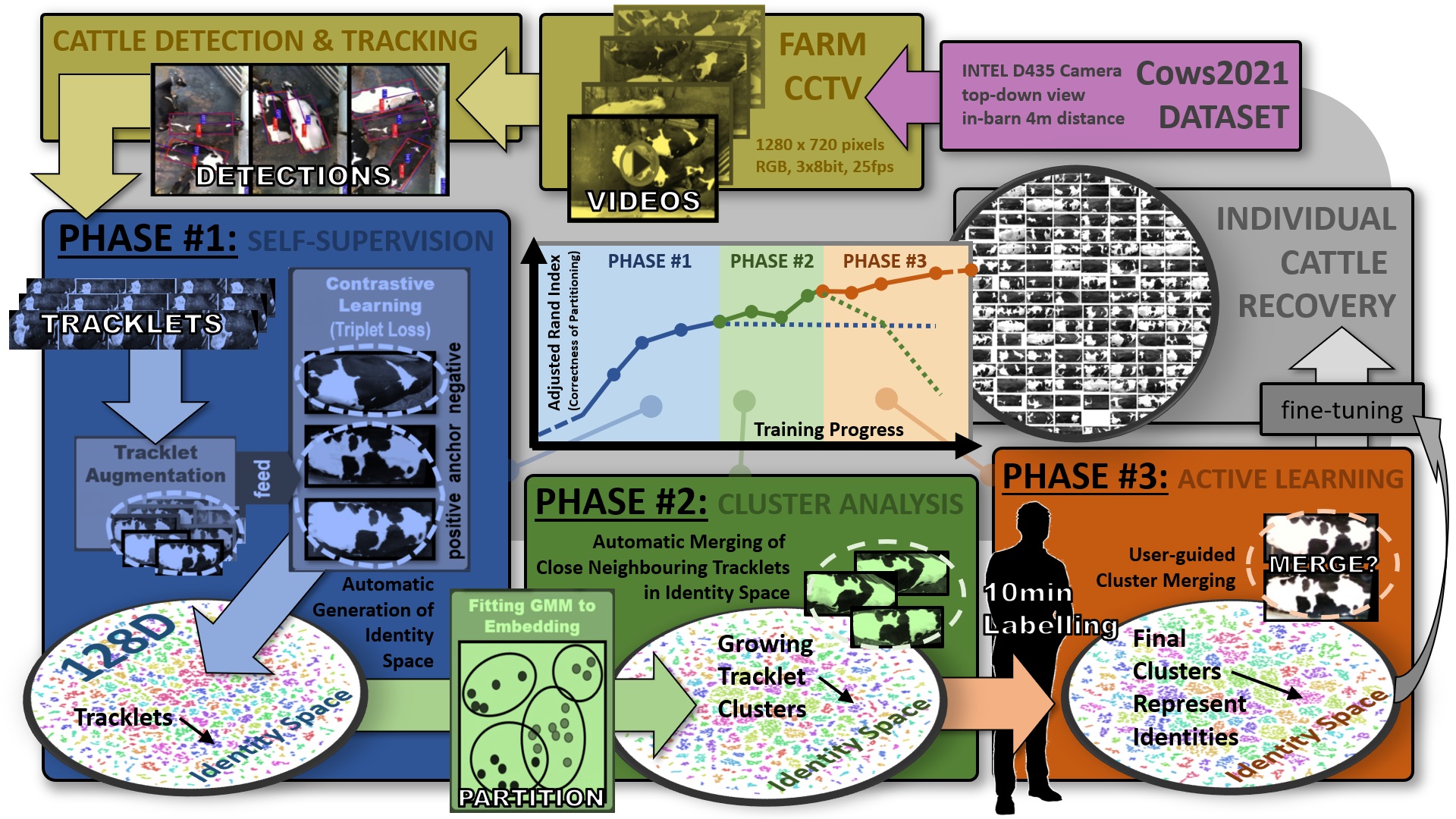} \vspace{-16pt}
    \caption{\small{$\bf{Conceptual}$ $\bf{Overview}$. \nwc{Using the public Cows2021 dataset~\emph{(purple)} we put forward a 3-phase identity learning process~\emph{(blue-green-orange)} using Holstein-Friesian detection sequences~\emph{(yellow)} in RGB videos from an operating farm. Our approach combines self-supervised metric learning~\emph{(blue)}, cluster analysis~\emph{(green)}, and active learning~\emph{(orange)} to iteratively improve performance~\emph{(middle plot)} under a minimal labelling regime. First, single animal tracklets are formed via automated oriented object recognition and tracking~\emph{(yellow)}. All frames from the tracklets are then mapped into a 128-dimensional identity space via metric learning and self-supervision~\emph{(blue)}. On this space we perform cluster analysis and merge tracklets to form growing clusters of individual cattle~\emph{(green)} used for fine-tuning. We identify the inflection point at which machine merging cannot compete with user input and thus determine when active labelling~\emph{(orange)} is needed. We track the evolving identity space via t-SNE and show significant improvements beyond self-supervision, reducing annotation requirements to around ten minutes.}
    %classification or clustering
    } 
    }\vspace{-13pt}
    \label{fig:overview}
\end{figure}
\vspace{-11pt}

\noindent{We will show that each of these phases can contribute to improving learning iteratively and together lead to benchmarks closer to fully supervised learning performance -- yet at a fraction of the labelling effort. Fig.~\ref{fig:overview} illustrates the proposed approach visually.}
 \ \\
  $\bf{Paper}$ $\bf{Contribution.}$ 
 \nwc{Overall, this paper makes three key contributions to the field of visual learning in precision farming:\vspace{-7pt}
 \begin{enumerate}
    \item A principled and practical three-phase hybrid deep learning framework with reduced labelling requirements for training coat-based biometric cattle ID systems.
    \item Integration of deep learning with a fast labelling approach (one-click same or different ID selection at 30 queries per minute) avoiding specific identity annotation, cold start issues~\cite{DBLP:conf/nips/KonyushkovaSF17}, or seed labelling requirements.
    \item A detailed system evaluation and comparative analysis on real-world farming videos across the public Cows2021 dataset including the identification of inflection points in the learning where human intervention becomes vital.
\end{enumerate}\vspace{-7pt}
We proceed by providing details on the dataset, implementation, experiments, and results. Finally, we discuss how these methods and insights can be utilised for more rapid ID system roll out in precision farming.}\vspace{-10pt}

\section{Dataset}
\begin{figure}[t]
    \centering
    \vspace{-13pt}
    \subfloat%[Tracklets from frames in videos]
    {\includegraphics[width=1\textwidth,
    trim= 0 5 0 5, clip]{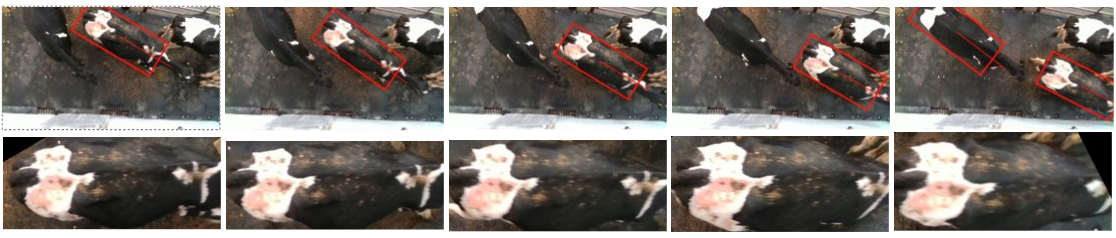} \vspace{-0pt}}
    \\
    \subfloat{\includegraphics[width=1\textwidth,
    trim= 0 5 0 5, clip]{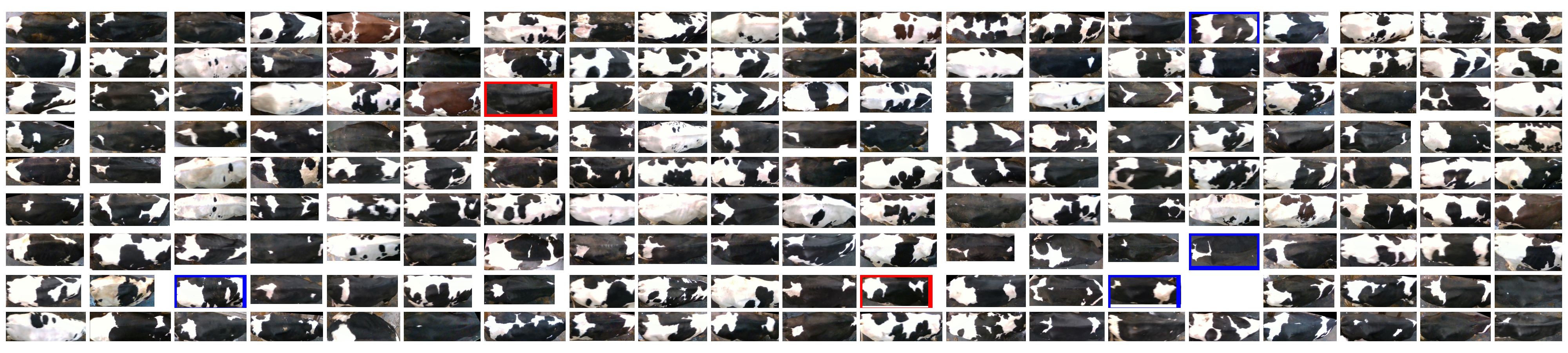} \vspace{-8pt}}
    \caption{\small{$\bf{Dataset}$. \nwc{Our training data contains 23,350 frames (4670 frames and their four augmentations) from 301 RGB videos resolved at $1280\times 720$ pixels. Automated oriented object detection and tracking~\cite{gao2021towards} of~these frames \textit{\textbf{(top)}} yields 435 normalised tracklets \textit{\textbf{(middle)}}. 
    \textit{\textbf{(bottom)}}
    The tracklets cover 155 individuals, all of which feature in the training set. 153 individuals are used in the test set (5344 extra images), and 149 in the validation set (2480 extra images). The four all-black cattle are represented here by a single image. Red boundaries denote individuals not included in the test set, both red and blue boundaries are not included in the validation set, and the last 24 individuals (654 images) after the blank area are unseen cattle during training used for open-set evaluation only.
    }}
    }\vspace{-16pt}
    \label{fig:data}
\end{figure}

\vspace{-10pt}
\nwc{The \textsl{Cows2021} dataset~\cite{Cows2021} 
contains 720p HD digital RGB video data from a working dairy farm in the United Kingdom. It was captured by a single-view top-down camera placed $4m$ above the ground between milking parlour and holding pens and operating at 25Hz (see Fig.~\ref{fig:data}{\textit{(top)}}).
Excluding three particular animal identities (IDs: 155, 169, 182) with too little data for effective verification options or poor quality images,
%\footnote{The pattern of the shoulder or bottom of cattle is missing as the cattle is at the edge of an image. We provide a subset of the test and validation data containing good-quality images. in the test and validation set}
we utilise 
%301 videos of 
the remaining herd covering 179 individual cattle (see Fig.~\ref{fig:data}{\textit{(bottom)}}) in our study. The dataset is first automatically processed into 435 tracklets from 301 videos produced by a deep object detection (see Fig.~\ref{fig:data}{\textit{(top)}}) and tracking framework for cattle detailed in~\cite{gao2021towards}. This ID-agnostic extraction performs rotational normalisation resulting in tracklets (see Fig.~\ref{fig:data}{\textit{(middle)}}) that contain exactly \textsl{one} individual with an average number of $1.45$ tracklets per video. Further statistics and data splits are given in Fig.~\ref{fig:data}. Note that four totally black cows (IDs: 54, 69, 73, 173) were treated as one individual during training and were excluded in the validation and test sets to avoid systematic errors arising by mixing this particular physical anomaly with other aspects of performance. Finally, for open-set testing we withheld 24 individual cattle from training and regular testing altogether. Since each tracklet contains data of only a single individual, self-supervision can be used to associate images to identity classes.}

% individual calculation: 186-1, 4in1 black (IDs: 54, 69, 73, 173) , 3 few  (IDs: 155, 169, 182.

\vspace{-10pt}
\section{Implementation}
\vspace{-7pt}
\subsection{PHASE \#1: Self-Supervision}
\label{Sec:Self}
\vspace{-5pt}
\subsubsection{Metric Deep Learning and Identity Space Construction.}
\begin{figure*}[t]
    \centering
    \vspace{-13pt}
    \subfloat[\nwc{START: Identity Space (ImageNet-Initialised)}]{\includegraphics[width=0.32\textwidth,trim=20 20 20 20, clip]{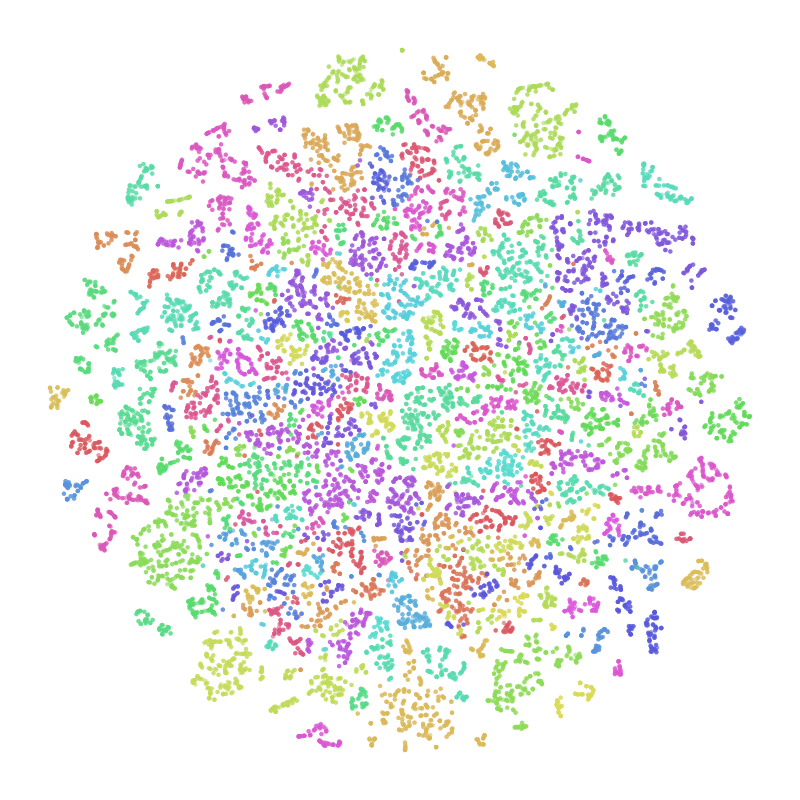}} % done#
    \hfill
    \subfloat[\nwc{PHASE \#1: Self-Super-vised Metric Learning}]{\includegraphics[width=0.32\textwidth,trim=20 20 20 20, clip]{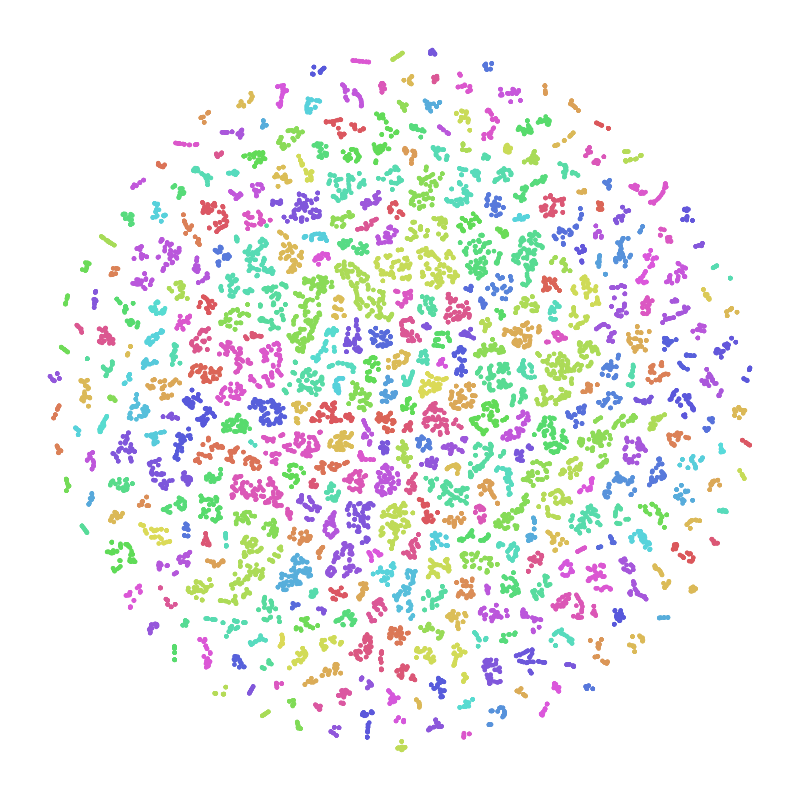}} %tsne_before
    %all_distance_0_9_first50c045_model_stateFolder_label on tsne.png}} 
    \hfill
    \subfloat[\nwc{PHASE \#2: Automatic Cluster Merging}]{\includegraphics[width=0.32\textwidth,trim=20 20 20 20, clip]{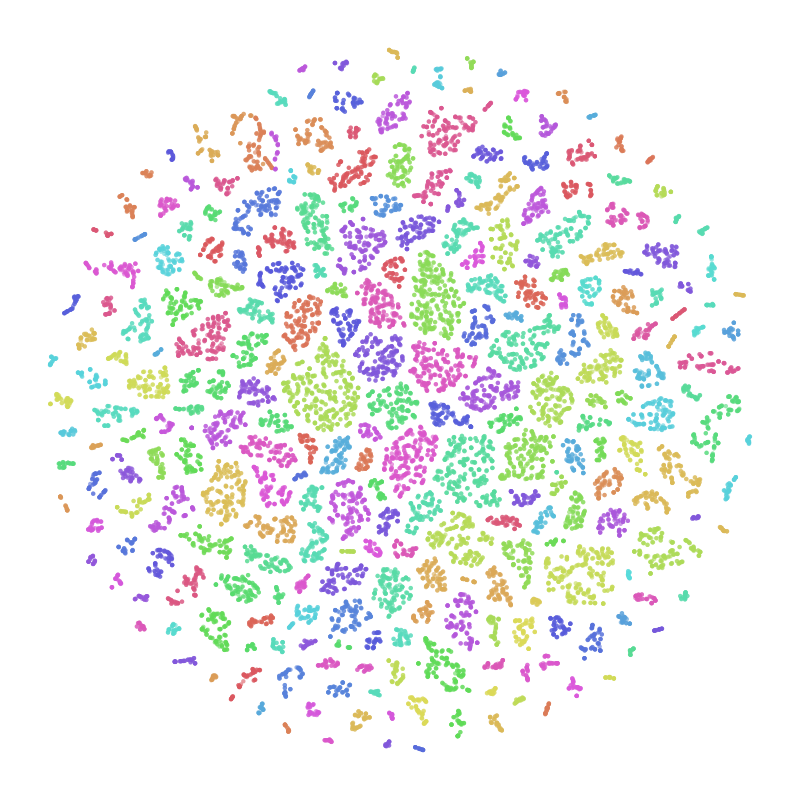}}  % 101 questions 
    % all_distance_1_3_first100c045_model_stateFolder_label on tsne
    \\
    \vspace{-1pt}
    \subfloat[\nwc{PHASE \#3: 2 Minutes Merging (Human Annotation)}]{\includegraphics[width=0.32\textwidth,trim=20 20 20 20, clip]{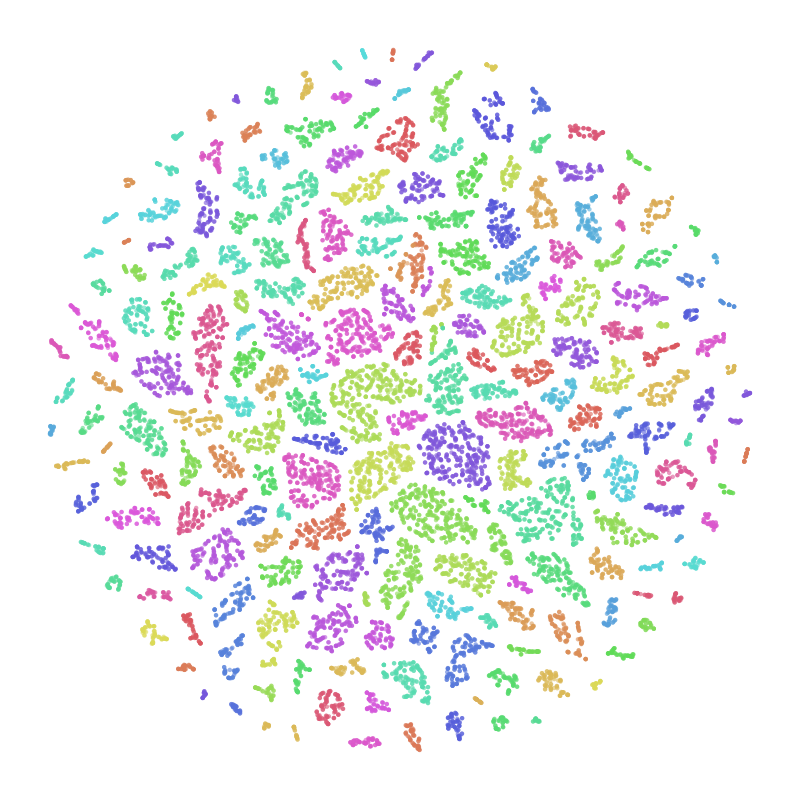}} 
    \hfill
    \subfloat[\nwc{PHASE \#3: 10 Minutes Merging (Human Annotation)}]{\includegraphics[width=0.32\textwidth,trim=20 20 20 20, clip]{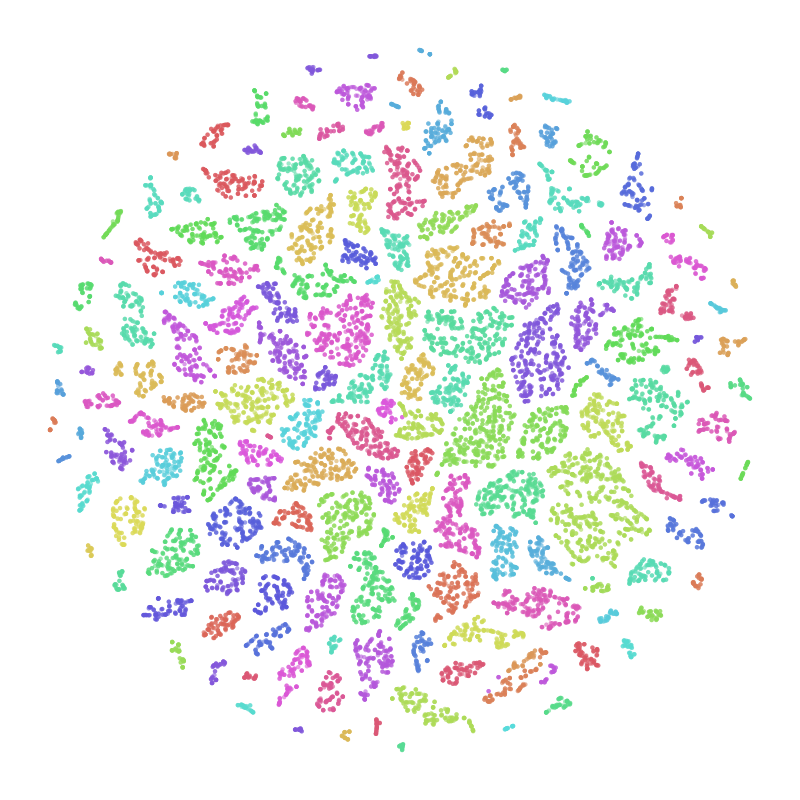}} %  405 questions  45x10 =450
    % all_distance_5_405qc020_model_stateFolde
    \hfill
    \subfloat[\nwc{Ideal Partitioning}]{\includegraphics[width=0.32\textwidth,trim=20 20 20 20, clip]{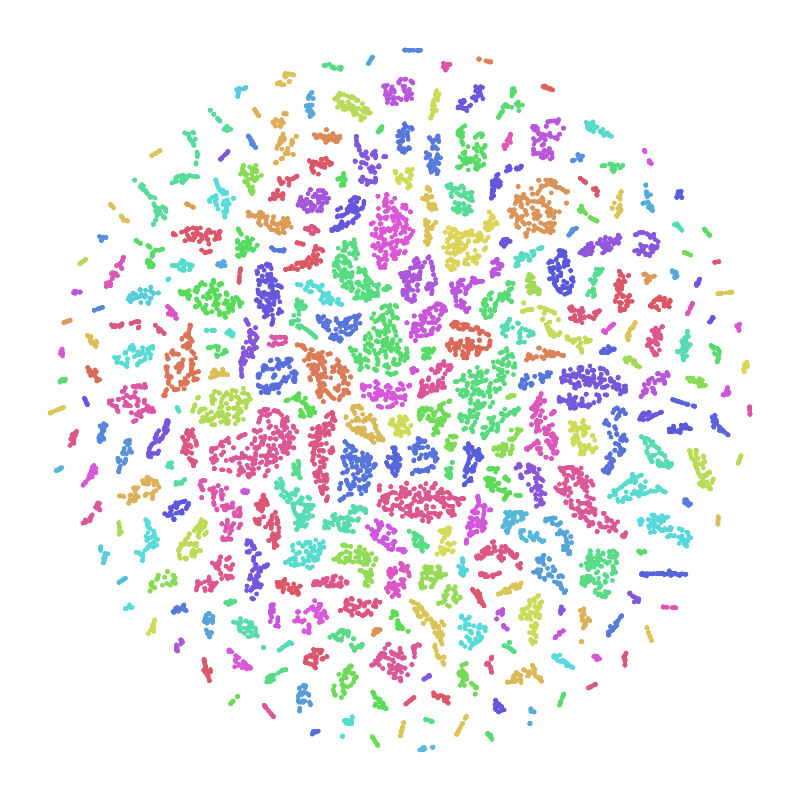}} \vspace{-5pt}
    %left bottom right top %width=0.35\textwidth,
    \caption{\nwc{\small{$\bf{Identity}$ $\bf{Space}$ $\bf{Evolution.}$ 2D t-SNE visualisations of the training data projected into the generated 128D latent identity space through the different phases of learning where colours relate to ground truth identities. Note that after \textit{\textbf{(a)}}~ImageNet initialisation, \textit{\textbf{(b)}}~self-supervision yields a tracklet-structured space where around three times more clusters are present than individual identities. \textit{\textbf{(c)}}-\textit{\textbf{(e)}}~Merging of these groupings and subsequent fine-tuning based on this leads to the step-wise discovery of actual identities and fewer and fewer overall clusters, first conducted via automatic cluster analysis in Phase \#2 and finally via active learning in Phase~\#3. }}}\vspace{-5pt}
    \vspace{-6pt}
    \label{fig:training_embeddings}
\end{figure*}

%--------------------------------------
\nwc{As our base architecture we use a ResNet50~\cite{he2016deep} backbone pre-trained on ImageNet~\cite{deng2009imagenet} modified to have a fully-connected final layer mapping to a latent $128$-dimensional vector from triplet image inputs detailed in~\cite{gao2021towards}. To initialise this space for identity information without any supervision we treat each tracklet as a unique class representing the same, unknown individual forming a set of `positive' image samples. One video may contain one or more tracklets. We pair these sets against `negative' samples from cattle shown in other tracklets from different videos, i.e. we use a video-aware sampling strategy. Statistically, the fact that the same `positive' individual may -- with a small chance -- appear in some `negative' sample in a different video is accepted as training noise during this initialisation stage. For self-supervised learning we use reciprocal triplet loss~($\mathbb{L}_{RTL} $)~\cite{masullo2019goes} leading to an initial version of the identity space that optimises:}\vspace{-7pt}
\begin{equation}
\label{eq:reciprocal-triplet-loss}
    \mathbb{L}_{RTL} = d(x_a, x_p) + d(x_a, x_n)^{-1}\vspace{-7pt}
    %\mathop{min}\limits_{\theta}{(d(x_a, x_\theta) + \frac{1}{d(x_a, x_n)}\vspace{-2pt)})}
\end{equation}
\nwc{where~$d$ denotes the Euclidean distance, $x_a$ and~$x_p$ are sampled from the `positive' set and~$x_n$ is a `negative' sample. We utilise online batch hard mining~\cite{hermans2017defense} expanded to a search that exploits both anchor and negative samples~(see code base for implementation details). Training of this stage took approx. 7 hours on an RTX2080 node using SGD~\cite{robbins1951stochastic} over $50$ epochs with batch size~$16$, learning rate~$1 \times 10^{-3}$, margin $\alpha=2$, and weight decay~$1 \times 10^{-4}$. The pocket algorithm~\cite{stephen1990perceptron} against the validation set was used to address network overfitting according to the Adjusted Rand Index (ARI). Using t-distributed Stochastic Neighbour Embedding~(t-SNE)~\cite{van2008visualizing} for visualisation, Fig.~\ref{fig:training_embeddings}{\textit{(a)}}) depicts the initial identity space after ImageNet pre-training and before self-supervision. As described above, the self-supervised metric learning then initialises this identity space as given in Fig.~\ref{fig:training_embeddings}{\textit{(b)}} leading to distinct local groupings of data points that relate to tracklets (i.e. image groups) of single cattle identities.} 
%Constrained self-training 

%---------------------------
% fig log:
% c: all_distance_2_1_173q_selfc032_model_state
% 228-173= 55 \30 = 2 min     d
% 277-173= 105 \30 = 3.5 min  
% 408-173= 235 /30 = 8  min   
% 471    = 298 /30 = 10  min  e

% d: FEB/self_after173_3_9_53q/1-1_92-96/3visual/c054_model_state
%self_after173_3_9_53qc054_model_state
% e  FEB/self_after173_10_298q/1-1_92-96/3visual/c058_model_state
% self_after173_10_298qc058_model_state
%---------------------------

\vspace{-10pt}
\subsection{PHASE \#2: Cluster Analysis}
\label{Sec:clus}
\vspace{-5pt}
\subsubsection{Clustering the Identity Space.} 
\label{Sec:aid}
\begin{figure}[t]
\vspace{-13pt}
    \includegraphics[
    width=125pt,height=125pt]{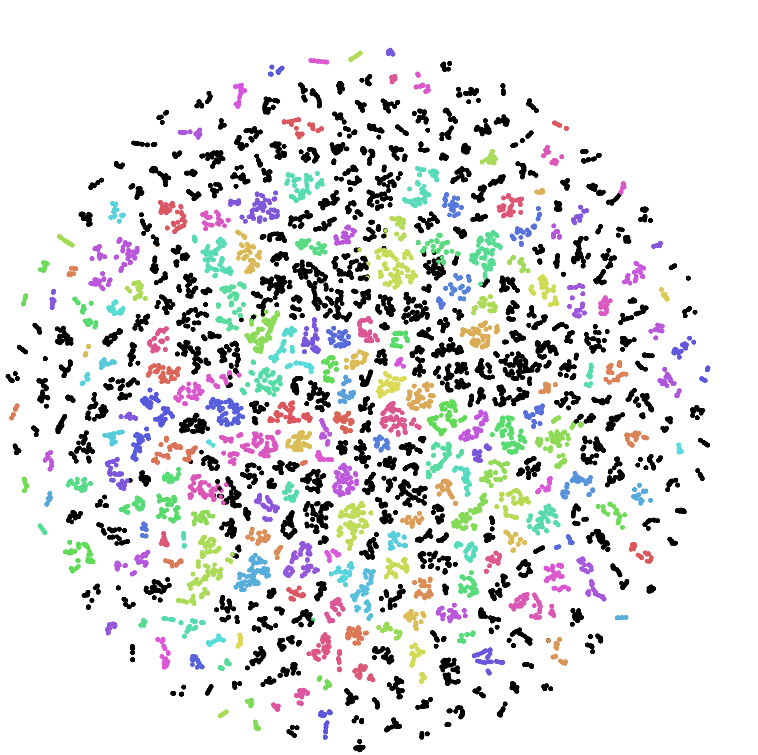}
    \includegraphics[
    width=220pt,height=115pt]{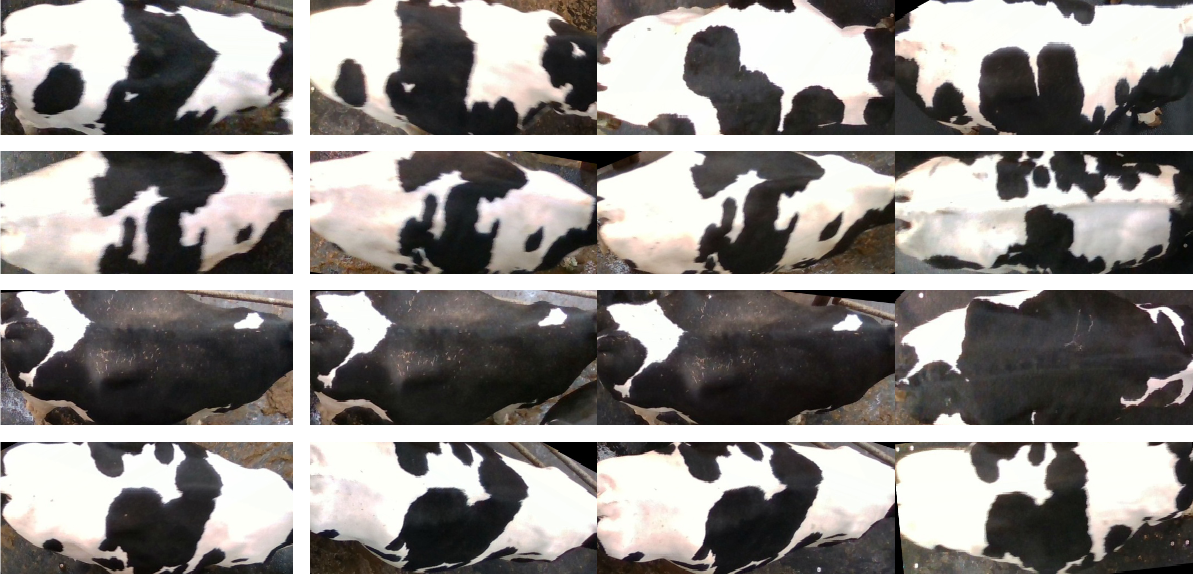}
    \vspace{-15pt}
    \caption{\small{$\bf{GMM}$ $\bf{Clustering}$ $\bf{and}$ $\bf{Tracklet}$ $\bf{Ranking}$.} \nwc{\textit{\textbf{(left)}} Training data of identities \emph{(colours)} assigned to the ground truth correctly and \emph{(black)} mismatched projected into the identity space after self-supervision with GMM. \textit{\textbf{(right)}} Four examples \emph{(rows)} of sample images from \emph{(leftmost)} anchor tracklets~(with ID~$\tau_m$) and the top three similar tracklets with IDs $\tau_{l^{m}_{2}}$, $\tau_{l^{m}_{3}}$ and $\tau_{l^{m}_{4}}$ forming candidate tracklet ID pairs $(\tau_m, \tau_{l^{m}_{k}})\in R$  to be merged as described in Algorithm~\ref{alg:GMM2}.}
    }
    \vspace{-16pt}
\label{fig:candidates}
\end{figure}

%--------------------------------------
\nwc{Our aim is that the identity space should be partitioned such that all tracklets are grouped to relate to individual cattle identities. However, we start by discovering clusters as given by the data in the space so far. We achieve such a partitioning by fitting a Gaussian Mixture Model~(GMM)~\cite{reynolds2009gaussian,scikit-learn} with $g$ = 450 components (this parameter is non-critical, provided that is is set to a number a little over the total number of tracklets) over 150 iterations to the tracklet-initialised identity space~(see Fig.\ref{fig:candidates}{\textit{(left)}}). Our approach is to utilise the discovered clusters to merge tracklets automatically into groups that represent the same individuals.}

\vspace{-10pt}
%-------------------------------------------
\subsubsection{Associating GMM Clusters with Tracklets.} 
\label{Sec:sample}
%-------------------------------------------
\nwc{We hypothesise that different GMM clusters which cover tracklets of the same individual should be near to each other in identity space and that the similarity between tracklets can be evaluated based on Euclidean tracklet-to-tracklet distances.
Therefore, we avoid calculating the distance of all the combinations of tracklet pairs and instead work with the top-$k$ matches to each tracklet utilising the GMM for calculation. The pseudocode is presented in Algorithm~\ref{alg:GMM2}. First, all data points $t\in T$ in the 128-dimensional embedding space are partitioned via the fitted GMM yielding $g=450$ data point clusters (i.e. GMM components) $T^{m}\subset T$ where $m\in [1,g]$ denotes the cluster label.  Next, the dominant tracklet ID marked as $\tau_{m}$ is assigned to each cluster label $m$. Let $T_{\theta}\subset T$ be the set of datapoints associated with the input tracklet ID $\theta$. The intersection set $T_{\theta}\bigcap T^{m}$ denoted as $T^{m}_{\theta}$ is then used to find the most frequent tracklet ID $\theta$ therein denoted $\tau_m$ and derived via a function `Mode' (line 8 of the Algorithm).}
% \newpage
\vspace{-10pt}
\subsubsection{Selecting Candidates for Merging.} 
\nwc{Next, we find candidate tracklet pairs for merging based on the associated GMM information. We use the fact that the GMM can return probabilities that an input $t$ belongs to \emph{any} particular cluster $m$. 
Across all data points in any GMM cluster $T^{m}$ we collect the $k^{th} (k\in[2,4])$ most likely cluster labels, that is for each cluster $T^{m}$ and $k$ we find the most frequent cluster label $l^{m}_k$ via a function `Select', which effectively identifies the $k^{th}$ nearest cluster to $T^{m}$. Each cluster identified by $l^{m}_k$ has itself a dominant tracklet ID $\tau_{l^{m}_{k}}$. With this information in hand, we finally compile the list $R$ of candidate tracklet ID pairs by pairing the dominant tracklet ID $\tau_m$ with~$\tau_{l^{m}_{k}}$.}

\vspace{-4pt}
\begin{algorithm}[t]
  \SetAlgoLined  
  \KwIn{
   \white{..}Data Points $T$ in Identity Space;\\
   \white{...............}Data Points $T_\theta\subset T$ with tracklet ID $\theta$;\\
   }
  \KwOut{Candidate List $R$ of tracklet ID pairs;}
   $T^{m}$\ $\leftarrow$ GMM$_{g=450}$($T$), $T$ is partitioned into clusters $T^{m}$ with labels $m$;\\
   $T^{m}_\theta = T_{\theta}\bigcap T^{m}$, Data Points with tracklet ID $\theta$ in cluster $T^m$;\\
%   $c_{m}$ $\leftarrow$ $Label(T^{m})$, the label of cluster $m$\;
  \For{$k\leftarrow 1$ \KwTo $4$}
  {
      \eIf{$k = 1$}{
        $\tau_{m}$ \ $\leftarrow$ Mode ($T^{m}_{\theta}$);
       % $P_{\tau_{m}, m}$   $\leftarrow$ list.append (Match($\tau_{m}$, $m$)), $m$ in range(1,$g$)\;
      }{
        $l^{m}_{k}$\ \  $\leftarrow$ Select ($k$, GMM, $T^m$, $T$);\\
        %$c_{m}$  $\leftarrow$ Find $\tau_{m}$ in $P_{\tau_{m}, l^{m}_{k}}$,  \qquad\  \qquad\  \ \  $m$ in range(1,$g$)\;
        % $c_{m}$  $\leftarrow$ $\tau_{l^{m}_{k}}$,  \qquad\  \qquad\  \ \  $m$ in range(1,$g$)\;
         $\tau_{l^{m}_{k}}$  $\leftarrow$ Mode ($T^{l^{m}_{k}}_{\theta}$);\\ %where $T$ $\in$ $T^{m}$), \;
        $R$ \ \ \ $\leftarrow$ list.append ($\tau_{m}$, $\tau_{l^{m}_{k}}$);
      }
    }
  \caption{Find candidates of tracklet pairs to merge based on GMM fit} \label{alg:GMM2}
\end{algorithm}
\vspace{-10pt}
\subsubsection{Candidate Ranking and Cluster Merging.}
\label{Sec:Queries Ranking}
\nwc{We now rank the generated candidate tracklet pairs $(T_\theta,T_\eta)$ associated to the list of tracklet ID pairs in $R$ based on the tracklet-to-tracklet distance between each two tracklets as suggested in previous work~\cite{wang2018deep}. The distance between two tracklets is calculated as the mean Euclidean distance between all point pairs that exist between the tracklets.
Candidates ranked towards the top are considered closest in identity space and have indeed a high chance of representing the same individual. The dotted brown curve in Fig.~\ref{fig:result1}~\textit{(left)} confirms this hypothesis against the ground truth by tracking the rate of correct merging. The graph shows a clear downwards trend of this rate when moving lower down the ranking. The rate first declines slowly, but then drops~rapidly.}
\vspace{-12pt}
\subsubsection{Identity Space Update via Fine-Tuning.}
\label{Sec:Tune}\nwc{After merging, we fine-tune the network for a further 5h of training as described in Section~\ref{Sec:Self} in order to incorporate the new information derived from cluster merging. This yields an updated network which defines a new embedding function for the identity space. By measuring the validation performance of this network against the number of tracklet mergers (or queries) we identify the point at which automatic merging cannot improve network performance (measured via ARI) anymore. This occurs after 173 queries as shown as the peak of the green curve in Fig.~\ref{fig:result1}~\textit{(left)}. Automatic merging after this point is thus too noisy to further improve performance. Consequently, user interaction via active learning should start at this point if further performance improvements are required.}

\vspace{-10pt}
\subsection{PHASE \#3: Active Learning}\label{Sec:act}
\vspace{-5pt}
\subsubsection{User-guided Fine-Tuning.}
\begin{figure}[h]
    \centering
    \vspace{-13pt}
    \subfloat{\includegraphics[width=180pt,
    height=100pt,
    trim= 5 10 5 5, clip]{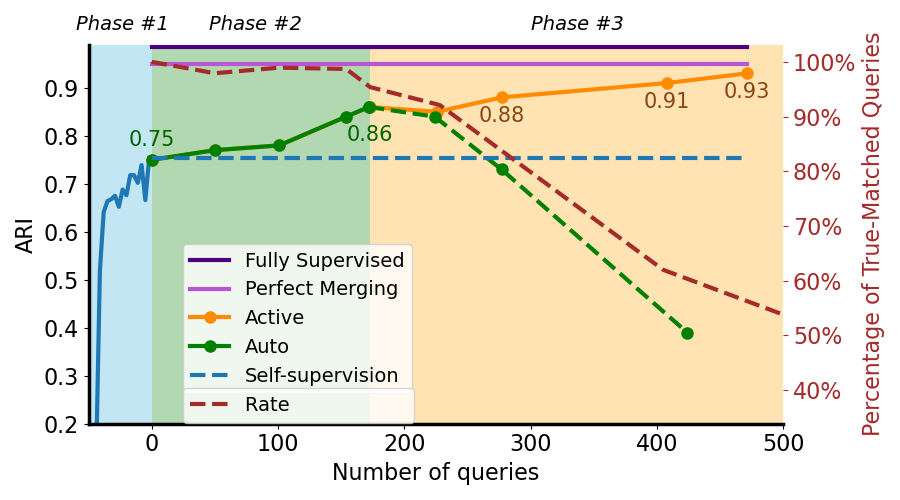}} %number_tr %left bottom right top 
    \hfill
    \subfloat{\includegraphics[width=165pt,
    height=100pt,
    trim= 5 10 5 5, clip]{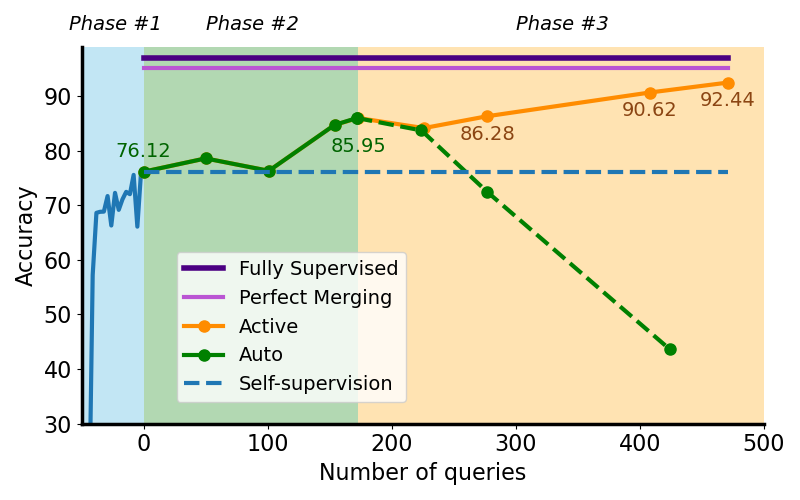}} \vspace{-4pt}
    \caption{\small{\blk{$\bf{Individual}$ $\bf{Identification}$ $\bf{Performance}$.} \nwc{System performance (closed-set) across the three learning phases shaded as blue, green, and orange. Benchmarks provide \textit{\textbf{(left)}} ARI measurements and \textit{\textbf{(right)}} Top-1 ID accuracy along evolving system training. Evaluation is conducted on 5k+ unseen test images. Phase \#1 in blue depicts results of self-supervised metric learning over the last 50 epochs leading up to the 0 mark of the abscissa indicating the start of tracklet merging. Phases \#2 and \#3 show results of fine-tuning after automated and manual merging, respectively. The abscissa tracks the number of mergers (or queries). Note, that at 173 queries, automatic merging~(green curve) peaks and further mergers are too noisy to improve performance without manual correctness checking via active learning~(orange curve). Note further that performance after~{10 min} of labelling at accuracy $92.44\%$ is close to the performance of perfectly merging \emph{all} clusters exhaustively~{($95.17\%$)} based on ground truth. The dotted brown curve on the left quantifies the proportion of correct merger suggestions.
    }}}
    \vspace{-18pt}
    \label{fig:result1}
\end{figure}

\nwc{Active learning~\cite{ren2021survey,settles2009active} aims to interactively annotate the most informative samples from the training model, followed by a model update. A domain-relevant sample selection is absolutely critical~\cite{lindenbaum2004selective,freytag2014selecting,wang2016cost}. For our case, having constructed a domain-informed ranked list~$R$ of potential tracklet ID mergers already, we will further utilise this information to guide active learning. Essentially, we will continue the tracklet merging strategy as before, but exploit user input to verify the merging instead of blindly accepting it. Thus, the final phase refines the identity space with a query sequence of one-click user inputs and utilises the 
%remaining ranked list~$R$ of potential tracklet mergers not used so far in automatic merging. 
new candidate ranking after another round of self-supervision retrained with the automatically merged tracklets.
For manual annotation, a human is shown the next highest ranking potential merger tracklet pair from $R$ presented as a single image from each of two tracklets. By confirming the identity to be the same or different only (one click), the human annotator throughput with this approach is extremely high at approximately 30 answers per minute. The approach avoids specific identity annotations (i.e. selection from a catalogue) and it has no cold start issues~\cite{DBLP:conf/nips/KonyushkovaSF17}. Practically, whenever two tracklets sharing the same identity are identified by an annotator, they are merged into a single tracklet and the number of tracklets is reduced as before during automatic merging. Note that it is possible that -- across various mergers -- more than two tracklets may merge into a single tracklet due to the transitive nature of the process. As before: after merging, {fine-tuning for 5 hours, is used to incorporate new information derived from cluster merging into the deep network}. The resulting networks are benchmarked as orange curves (marked `Active') in Fig.~\ref{fig:result1}.}

%====================================================================================
\vspace{-7pt}
\section{Experimental Results}
\vspace{-5pt}

%-------------------------------------------
%-------------------------------------------
\subsubsection{Closed Set Testing.}
\nwc{The proposed pipeline is first evaluated on the 5,344 unseen test images across 153 individuals. This is closed-set testing since all these individuals have been seen (in different videos) during training. A projection of the closed testset into the learned identity space is visualised via t-SNE in Fig.~\ref{fig:test}{\textit{(a)}}). In order to evaluate the clustering performance in identity space, we use two measures: the Top-N ID prediction accuracy and also the Adjusted Rand Index~(ARI)~\cite{hubert1985comparing} for assessing structural clustering similarity.}

\vspace{-14pt}
\subsubsection{Structural Clustering Benchmarks.}
\label{sec:ARI}
\nwc{We evaluate the structural similarity of the testset clustering produced by the trained network against the ground truth. We quantify the similarity via the ARI measure which operates on two data partitions and does not require any knowledge of ID labels. In particular, the testset clustering is derived by fitting a GMM (with the component cardinality of the testset) to the test data in the fine-tuned identity space. Fig.~\ref{fig:result1}~\textit{(left)} shows the evolution of ARI across the three learning phases. Self-supervision of Phase \#1 as described in Section~\ref{Sec:Self} leads to an initial ARI value of $0.75$\footnote{\nwc{Note that this performance improves on the self-supervision state-of-the-art~\cite{gao2021towards} in the domain by using the same network yet with our extended hard mining regime. Testing our Phase \#1 method on their testset improves ARI from their published ARI of $0.53$ to $0.65$.}}~(see blue curve). In Phase \#2, automatic tracklet merging and subsequent fine-tuning as described in Section~\ref{Sec:clus} steadily improves the ARI further to a peak value of $0.86$ at query 173~(see green curve). At this point, we find that any further automatic merging is not beneficial as depicted by the dotted green curve. In Phase \#3, user input is now utilised. The ARI can be increased further to $0.88$ after {3.5} minutes of human labelling, $0.91$ after {8} min, and 0.93 after 10 min. Perfectly merging {all} clusters exhaustively only leads to a small increase of ARI at~{$0.95$}.}
% 66.7 - 0.65
% 57.0 - 0.53
% 277-173= 105 \30 = 3.5 min  
% 408-173= 235 /30 = 8  min   
% 471    = 298 /30 = 10  min 

% \blk{
% With 154, 173 and 228 queries in the `Active Learning' curve, ARI reached 0.843, 0.852 and 0.883, separately. They are greatly outperforms the random sampling methods. Although more queries like 405 or 695 can be answered, the improvement is not as obvious as the former.}

%-------------------------------------------
\vspace{-14pt}
\subsubsection{Accuracy Stipulation.}
\nwc{Next, we evaluate identification accuracy against ground truth ID labels of the testset. In order to obtain IDs for our network output, each fitted GMM cluster used for structural benchmarking is assigned to the one individual ID having the highest overlap ratio with the ground truth, defined as: }
\vspace{-6pt}
\begin{equation}
\label{eq:overlap}
\mathbb{O}_{l}=C / L\vspace{-2pt}
\end{equation}
\nwc{where $C$ is the number of images in a GMM cluster that belongs to an individual, and $L$ is the total number of images of that individual. This produces the (GMM Cluster)-(ID Label) pairs required for accuracy evaluation against the ground truth.} 
% The relation between accuracy and merge annotation ratio is illustrated in Fig.~\ref{fig:merge}.

\begin{figure}[t]
    \begin{center}
    \vspace{-5pt}
    \includegraphics[height=108pt, %width=0.42\textwidth, 
    trim= 10 10 10 10, clip]{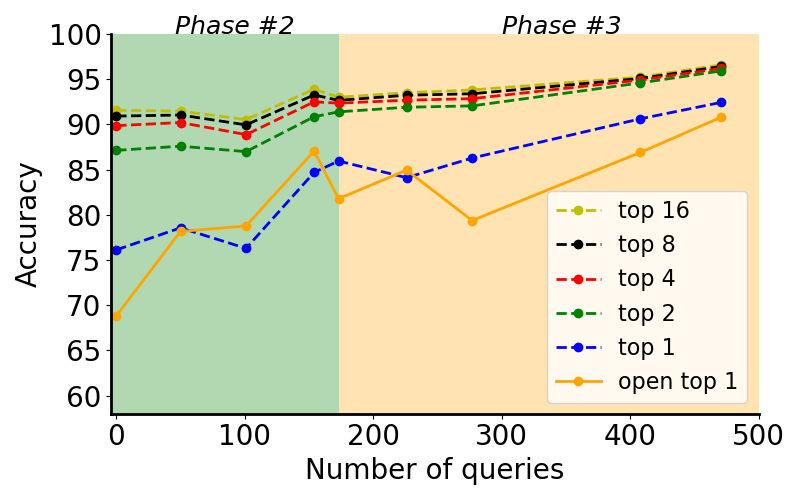}\white{...} \includegraphics[width=160pt, trim= 0 0 0 0, clip]{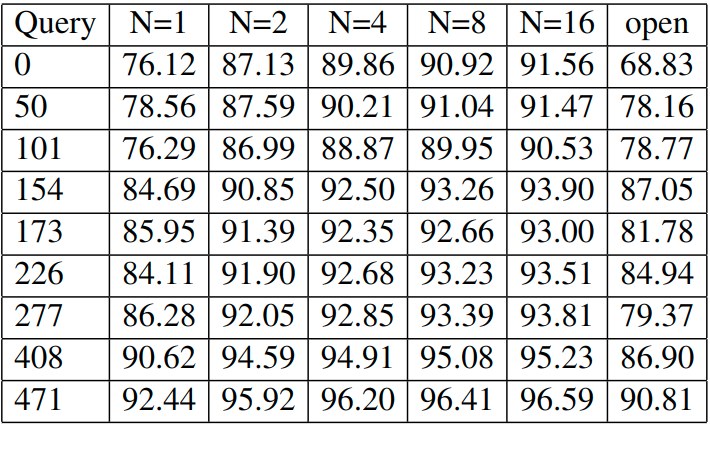} 
    \end{center}
    \vspace{-16pt}
    \caption{\small{{$\bf{Top}$-$\bf{N}$ $\bf{and}$ $\bf{Open}$-$\bf{Set}$ $\bf{Performance.}$} \nwc{Depicted is the Top-N identification accuracy of 153 individuals across the final two learning phases shaded green and orange, respectively. Benchmarks provide accuracy curves for Top-N closed-set (24 individuals) accuracy (detailed also as values on the \textit{\textbf{(right)}}) and Top-1 open-set accuracy in yellow. Note that open-set performance benefits strongly from the proposed Phase \#2 and \#3, leading to near closed-set performance.}
    } 
    }\vspace{-16pt}
\label{fig:topn}
\end{figure}

\vspace{-14pt}
\subsubsection{Top-1 Accuracy Benchmarks.} \nwc{The resulting Top-1 test accuracy for individual identification over the three training phases is given in Fig.~\ref{fig:result1}~\textit{(right)}. Generally, the plot structure is consistent with our ARI analysis in Fig.~\ref{fig:result1}~\textit{(left)} showing the effectiveness of all three phases of our approach across the two different performance measures. Phase \#1 leads to an accuracy of $76.12\%$ already on the test set, emphasising the strength of a tracklet-driven self-supervision signal during metric learning. In Phase \#2, automatic tracklet merging can further improve accuracy to a peak value of $85.95\%$~(see green curve). Note that automatic merging is imperfect and accuracy is not optimised directly by fine-tuning, thus a temporary decrease of accuracy is possible. In Phase \#3, human labelling allows a further increase in accuracy to $86.28\%$ after {3.5} minutes of human labelling, $90.62\%$ after {8} minutes, and finally $92.44\%$ after 10 minutes.}
% 277-173= 105 \30 = 3.5 min  
% 408-173= 235 /30 = 8  min   
% 471    = 298 /30 = 10  min 

\vspace{-14pt}
\subsubsection{Top-N Accuracy Benchmarks.}\nwc{For each GMM cluster assigned to an ID as per Eq.~\ref{eq:overlap}, one can alternatively rank all identities according to~$\mathbb{O}_{l}$where identities that have a~$\mathbb{O}_{l}=0$ form the tail of the ranking sequence with randomly assigned, remaining IDs. This process allows the creation of a Top-N~\cite{NIPS2012_4824} accuracy benchmark as shown in Fig.~\ref{fig:topn}. In this statistic, a prediction is considered correct if and only if the ground truth ID is found amongst the top N ranked IDs. As expected, for $N=1$ this equates to the traditional definition of accuracy. Fig.~\ref{fig:topn} emphasises that across Phases \#2 and \#3 improve accuracy across all N. This setting is particularly interesting if semi-automatic identification is used to present a system user with a set of N candidate identities for a query. For $N=16$ this setting leads to a near-perfect accuracy benchmark of~$96.59\%$ using only 10 minutes of labelling.}

%-------------------------------------------
\vspace{-14pt}
\subsubsection{Open-Set Accuracy Benchmarks.}
\begin{figure}[t]
    \begin{center}
    \vspace{-13pt}
          \subfloat[\nwc{CLOSED TESTSET, 153 animals. (10min Annotation)}]{\includegraphics[width=0.32\textwidth,trim=20 20 20 20, clip]{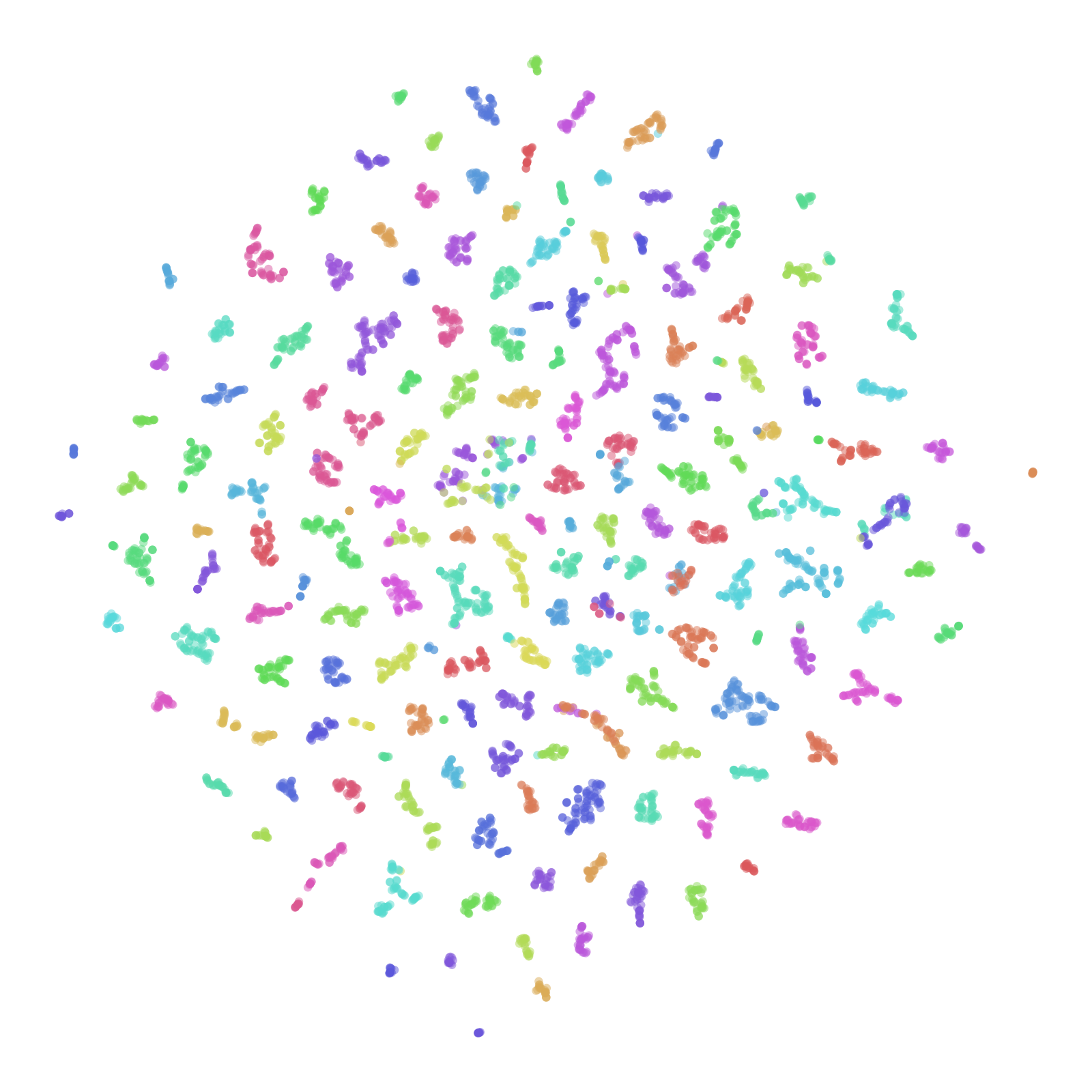}} % done#
    \hfill
    % \subfloat[\nwc{CLOSED TESTSET, 153 animals. (Fully Supervised)}]{\includegraphics[width=0.32\textwidth,trim=20 20 20 20, clip]{images/will_gmm.png}} 
    \subfloat[\nwc{CLOSED TESTSET, 153 animals. (Fully Supervised)}]{\includegraphics[width=0.32\textwidth,trim=20 20 20 20, clip]{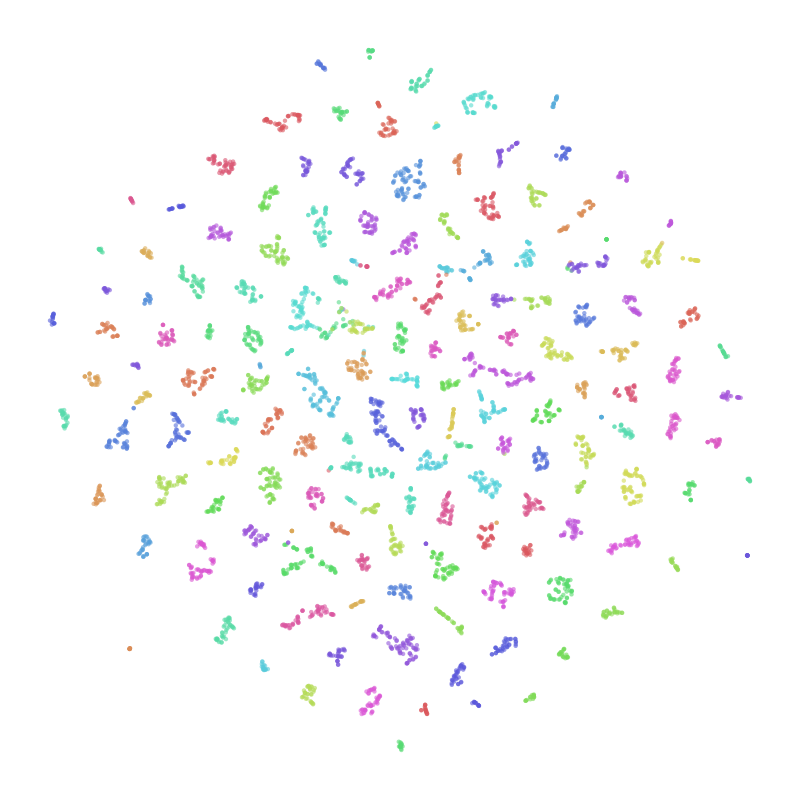}} 
    \hfill
    \subfloat[\nwc{OPEN TESTSET, 24 animals. (10min Annotation)}]{\includegraphics[width=0.32\textwidth,trim=20 20 20 20, clip]{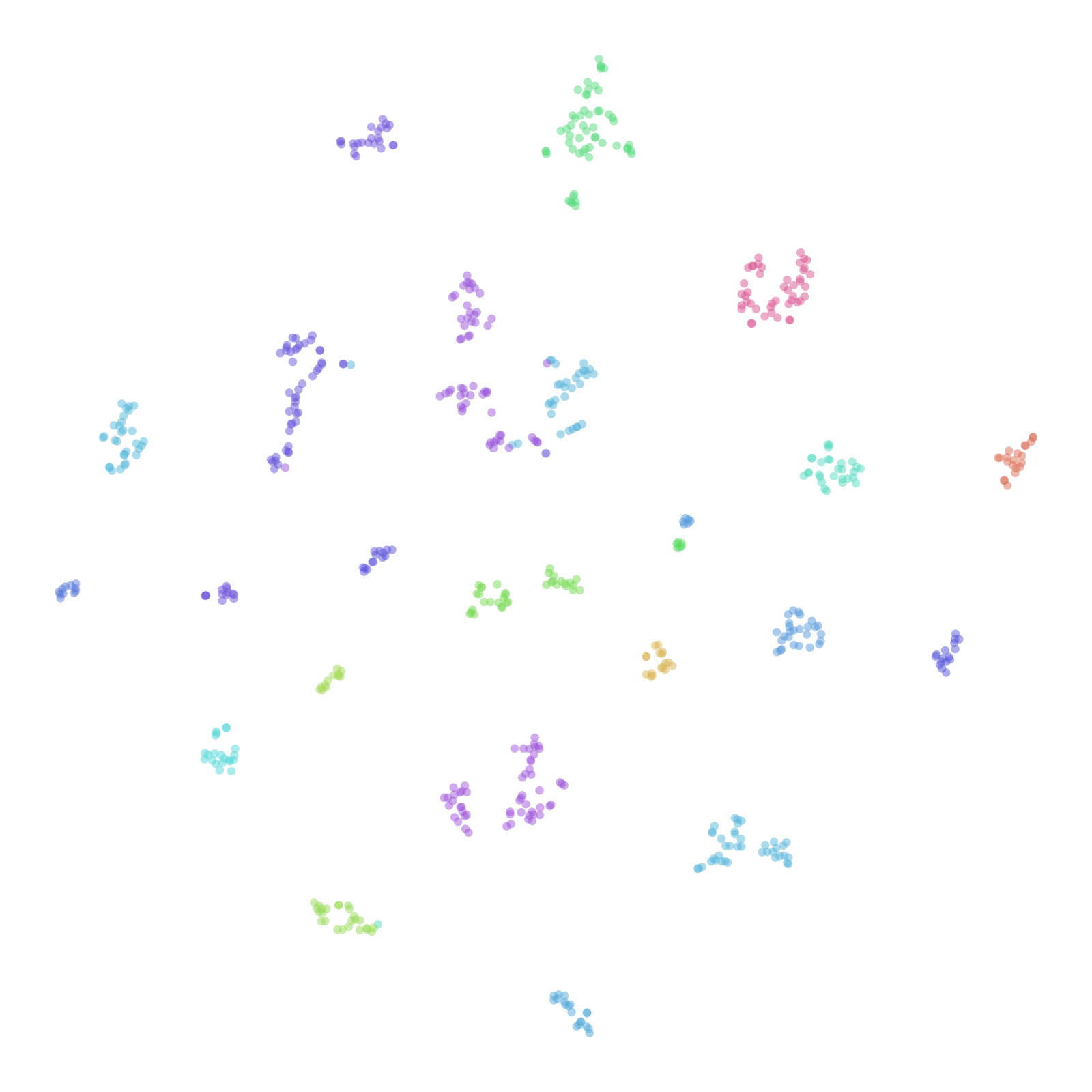}}  % 101 questions 
    % all_distance_1_3_first100c045_model_stateFolder_label on tsne
    \end{center}
    \vspace{-13pt}
    \caption{\small{$\bf{t}$-$\bf{SNE}$ $\bf{Testset}$ $\bf{Embeddings.}$ 
    } 
    \nwc{2D t-SNE visualisation of the closed and open test sets when projected into the identity space. \textbf{\textit{(a)}} shows the result for the proposed method at accuracy~{$92.44\%$} after all three stages of training and 10min of manual annotation. \textbf{\textit{(b)}} For comparison, we show the same testset projected into an identity space built from fully manual annotation at the Top-1 %accuracy~{$97\%$} 
    accuracy~{$96.05\%$} 
    using weeks of frame-by-frame identity labelling. \textbf{\textit{(c)}} When projecting unseen images of unseen animals into the space used in \textbf{\textit{(a)}} identities still cluster well (accuracy~{$90.81\%$}) confirming generalisation to a domain wider than the herd trained on.}
    }\vspace{-15pt}
\label{fig:test}
\end{figure}

\nwc{The proposed pipeline is secondly evaluated on the 654 unseen test images of 24 never seen individuals. A projection of this open testset into the learned identity space is visualised via t-SNE in Fig.~\ref{fig:test}~\textit{(c)}. This `Open-Set' testing is critical for stipulating how far performance can translate across herds or farms in a zero-shot identification paradigm. Fig.~\ref{fig:topn} depicts Top-1 open-set test performance across the final two phases of training. 
%{It can be seen that open-set performance lags behind closed-set performance consistently, however, the margin (gap between blue and yellow curves) is not large at $1.53\%$ after 10 min of manual labelling.}
It can be seen that open-set performance generally lags behind closed-set performance.
However, the margin (gap between blue and yellow curves)
is only $1.53\%$ after 10 minutes of manual labelling. 
This is promising, showing that the identity space created generalises well across the domain of Holstein-Friesians and not just the training herd.
}

%-------------------------------------------
        % Queries & 50 & 100 & 154 & 173  & 228 & 405 & 695 \\ 
        % ARI & 0.689 & 0.727 & 0.784 & 0.847 & 0.836 & 0.871 & 0.895\\ 

%------------------------------------------------------------------------

%--------------------------------------

%\begin{figure}[t]
 %   \begin{center}
  %  \vspace{-13pt}
  %  \subfloat{\includegraphics[height=100pt, %width=0.42\textwidth, 
  %  trim= 0 10 0 0, clip]{images/distance_distribution.png} %left bottom right top %width=0.35\textwidth,
%    }
 %   \hfill
  %  \subfloat{\includegraphics[height=100pt, %width=0.42\textwidth, 
  %  trim= 0 5 0 0, clip]{images/error.png} %left bottom right top %width=0.35\textwidth,

%    }
        %left bottom right top 
 
 %   \end{center}
  %  \vspace{-10pt}
  %  \caption{\small{\tmp{ $\bf{Distance}$ $\bf{Distribution}$ $\bf{between}$ $\bf{Tracklets.}$ from active learning.} $(right)$ queries vs the percentage of false-matched pairs with the ascending order of tracklet-to-tracklet distance. these are also the highest confident queries. Once the distance is over 2.2, true-matched pairs are rare.
   % } 
%    }\vspace{-16pt}
%\label{fig:distance_distribution}
%\end{figure}

%------------------------------------------------------------------------

%====================================================================================
\vspace{-7pt}
\section{Conclusion}
\vspace{-10pt}
%----------------------------------------------
\nwc{We have put forward a practical, three-phase deep learning approach for training an ID system for Holstein-Friesians on a farm, requiring only ten minutes of labelling. We showed that automatic identification of individual cattle in real-world farm CCTV can be achieved effectively by combining self-supervision, metric learning, cluster analysis, and active learning. We provided detailed explanations and key source code\footnote{\url{https://github.com/Wormgit/LabelaHerdinMinutes}} for full result reproduction and evaluated the approach using the publicly available Cows2021 dataset. Self-supervised metric learning was first leveraged to initialise an identity space where tracklets are considered a distinct entity. Grouping entities is then performed by automated merging via cluster analysis and active learning feeding into fine-tuning. Experimental results showed that cluster analysis and a few minutes of labelling after automated self-supervision can indeed improve the identification accuracy of closed and open test sets compared to self-supervision only. Despite superior performance of fully supervised systems which required weeks of frame-by-frame labelling in the past, our 10 minute labelling approach shows promising results indicating that human and machine reasoning in tandem can be integrated into visual cattle ID pipelines in a highly effective fashion requiring only minimal labelling effort.}

\clearpage
\bibliographystyle{splncs04}
\bibliography{egbib}

\end{document}